# An Evolutionary Algorithm for Error-Driven Learning via Reinforcement

Yanping Liu
*Key Laboratory of Behavioral Science*
*Institute of Psychology, Chinese Academy of Sciences*

&

Erik D. Reichle
*School of Psychology, University of Southampton, UK.*

*Note*: This is a draft; please do not cite without permission.

Address correspondence to:

Yanping Liu,
16 Lincui Road, Key Laboratory of Behavioral Science,
Institute of Psychology,
Chinese Academy of Sciences, Beijing, China.
Email: liuyp@psych.ac.cn.

Although different learning systems are coordinated to afford complex behavior, little is known about how this occurs. This article describes a theoretical framework that specifies how complex behaviors that might be thought to require error-driven learning might instead be acquired through simple reinforcement. This framework includes specific assumptions about the mechanisms that contribute to the evolution of (artificial) neural networks to generate topologies that allow the networks to learn large-scale complex problems using only information about the quality of their performance. The practical and theoretical implications of the framework are discussed, as are possible biological analogs of the approach.

Modern cognitive science has dispelled the long-held belief that the behavior of higher organisms is governed by a unitary learning system (e.g., Skinner, 1938) and has instead shown that behavior is controlled by several different learning mechanisms, each operating according to its own principles (Doya, 1999, 2000; Squire, 1993). With this progress, however, have come new questions about how these learning systems are coordinated so that organisms can learn the complex behaviors that are necessary to live and reproduce (e.g., see McClelland, McNaughton, & O'Reilly, 1995). The goal of this article is to address this question by considering two specific types of learning that have been widely studied and that are known to play critical roles in complex behavior: reinforcement learning and error-driven learning.

*Reinforcement learning* refers to a type of learning in which an organism's behavior is gradually shaped through contingencies of reward and punishment. In other words, the organism performs actions that result in rewards and/or punishments and then, over time, learns to perform those actions that maximize/minimize the amount of reward/punishment obtained from those actions and from the states that result from those actions. By learning to associate reward/punishment values with specific actions and their resulting states, the organism essentially learns which actions are most appropriate from each possible state. When viewed in this way, reinforcement learning refers to a general class of learning algorithms that has been extensively studied in psychology and that includes both classical (Pavlov, 1927) and operant (Thorndike, 1901) conditioning. And in the field of machine learning, reinforcement learning refers to a class of algorithms that—like their biological analogs—allow artificial systems to learn how to optimize their performance in various tasks (e.g., navigating through mazes; Sutton & Barto, 1998) by associating values with various actions and states. However, because reinforcement learning is contingent upon a one-dimensional error signal, its applicability to complex high-dimensional (e.g., non-linear) tasks has been severely limited (Sutton & Barto, 1998), as has its application to psychology (Gershman, Cohen, & Niv, 2010).

In contrast to reinforcement learning, *error-driven learning* refers to a type of learning in which the organism uses information about how its behavior differs from some "target" or desired behavior to modify its actual behavior so that the latter comes to resemble the former (McClelland et al., 1995; Rumelhart, Hinton, & Willams, 1986). Because error-driven learning uses the difference or *error* between the actual and desired behavior, and because this error can vary along multiple dimensions, error-driven learning is not limited like reinforcement learning and can be used to learn complex, high-dimensional problems (e.g., learning English spelling-to-sound correspondences; Seidenberg & McClelland, 1989). But this enhanced learning requires an explicit error or *teaching signal* that has often undermined its applicability in both psychology (Dayan & Niv 2008; Niv, 2009; Sutton, 1991) and machine learning (Alonso & Mondragón, 2005; Ye, Yung, & Wang, 2003). The strengths and weaknesses of error-driven learning thus complement those of reinforcement learning, and neither class of learning algorithms is sufficient to learn complex behaviors in the absence of an explicit "teacher" that provides information about how the behavior that is currently being performed differs from the behavior that must eventually be learned. Of course, the obvious way to redress this limitation is to somehow combine the two types of learning and thereby allow error-driven learning via reinforcement. We suspect that this "solution" was the one arrived at by evolution through natural selection, and in the remainder of this article we will describe an evolutionary-based algorithm that does exactly that—it allows error-driven learning to occur via simple reinforcement. As will be discussed below, this new algorithm: (1) allows artificial systems to learn problems through reinforcement that have heretofore been too complex to acquire, and (2) provides a hypothesis about the functional roles played by both error-driven and reinforcement learning in the acquisition of complex behaviors by real biological organisms.

**1. General Overview**

In biological organisms, the processes that support all types of learning are instantiated within networks of highly interconnected neurons. These *neural networks* receive input from the perceptual systems and then propagate this activity as output to motor effectors, allowing the organism to both perceive and interact with the physical environment. Learning in such a network thus occurs through the modification of the connection strengths among the neurons. For example, a one-layer network must learn the correspondences the between stimuli and responses by adjusting its connection strengths so that it can execute an appropriate response to each stimulus. By modifying its connection strengths in this manner, a network retains information in the structure that mediates the "mappings" between stimulus inputs and response outputs. However, it is important to note that this structure or *topology* is derived from two sources—the *macrostructure* of the network (e.g., the number of neurons, their pattern of connectivity to the perceptual and motor systems, etc.) that is presumably acquired very slowly through evolution, and the *microstructure* of the network (e.g., the strengths of individual connections) that is acquired more rapidly through learning. Both types of structure are therefore important determinants of the types of behaviors that a neural network is capable of learning.

The manner in which topology constrains learning has been conclusively demonstrated using the machine-learning analogs of the biological neural networks (e.g., for discussions of these constraints, see McClelland & Rumelhart and the PDP Resarch Group, 1986; Rumelhart & McClelland and the PDP Research Group, 1986). In these *artificial neural networks*, the connection strengths between "neurons" are modified using a variety of different algorithms, with perhaps the most widely used of these algorithms being *back-propagation* (Rumelhart et al., 1986). The primary virtues of back-propagation are its relative simplicity and the fact that it is sufficient to train multi-layer networks (i.e., networks in which input units are separated from output units by one or more layers of intermediate "hidden" units). Such multi-layered networks are necessary to learn tasks that require non-linear mappings between input and output, such as the classic *exclusive-or problem*, where either of two options results to one response but neither or both of the options results in the opposite response (Rumelhart et al., 1986). The main limitation of back-propagation is that it is a form of error-driven learning, and as such requires an explicit teaching signal during training. As already indicated, this restricts back-propagation to certain classes of problems (i.e., problems in which there is a teaching signal) and thereby undermines its possible utility in learning behaviors in which the only feedback is whether those behaviors resulted in positive or negative outcomes—the types of behaviors that can be acquired via reinforcement learning.

This restriction is unfortunate because, in addition to offering some degree of biological plausibility, artificial neural networks convey one other advantage that would otherwise make them ideal for learning complex behavior—the graded nature of their processing allows the networks to approximate functions very effectively. For example, in contrast to standard reinforcement-learning algorithms (e.g., *value-iteration*; Sutton & Barto, 1998), where each state and its associated value are discretely represented (e.g., as a collection of values in a look-up table), networks represent information about all states and their values within a single set of interconnected processing units, so that the network is able to interpolate and extrapolate from its knowledge and thereby generalize from known states to novel states.

As will become evident below, our new algorithm to enable error-driven learning via reinforcement exploits this characteristic of neural networks by allowing information about states and their values to be stored in a single set of connections, thus allowing the algorithm to generalize from situations used during training to completely new situations. The essence of this approach is that it acknowledges the constraints that are imposed by a network's macro- *and* microstructures, and in so doing incorporates an evolutionary-based method for developing macrostructures that are specialized for certain types of learning (i.e., microstructures).  More specifically, this approach uses a genetic algorithm (Holland, 1975) to evolve artificial neural networks that are capable of error-driven learning via reinforcement.  We will now describe this approach in four sections, explaining how our hybrid algorithm: (1) selects the appropriate macrostructure that supports learning; (2) simulates the emergence of the microstructure that reflects learning; and (3) implements reinforcement learning.  The final section (4) will then provide an overview of how these three processes are coordinated to allow artificial neural networks to perform error-driven learning via reinforcement.

## 2. Macroscopic Evolution of the Network Architecture

Efficient biological evolution entails safeguards to protect any innovations of the phenotype that offer a selective advantage (in terms of ecological fitness) to the genotype, while simultaneously detecting homology between genotypes and minimizing the structural complexity of the genotype.  These safeguards maximize the efficiency of biological evolution by minimizing the overall size of the evolutionary "search space" that must be "traversed" to produce organisms that are fit enough to compete and survive in specific ecological niches.  Our method of generating network topologies uses an algorithm (*NeuroEvolution of Augmenting Topologies*, or *NEAT*; Stanley & Miikkulainene, 2002) that was specifically designed to instantiate these three efficiency strategies as follows: First, innovations in network topology are protected (so that they have a reasonable chance of propagating from one generation of networks to the next) by *speciation*, or the evolution of separate species that comprise distinct populations of networks that have unique topologies and that only reproduce within their own population.

Second, homologous genotypes can be identified by historical "markers" (i.e., identifiers that reflect each network's evolutionary history) to allow for efficient identification and alignment of the genomes during the crossover stage of reproduction. And finally, the structural complexity of the genotypes is minimized by beginning the process of evolution with the simplest possible network topology—single-layer networks containing only input and output units.

In our approach, each genome is a linear array of genes that represents each network's topology. There are two basic types of genes: (1) *node genes* that determine the functional role of each node (i.e., input, output, or hidden unit) in the network, and (2) *connection genes* that determine the patterns of connectivity among the nodes in the network. Figure 1 is a schematic diagram showing the relationship between a single example genome (i.e., the genotype) and the network that it instantiates (i.e., the phenotype). As indicated, each node gene indicates the functional role of a node and provides a unique identifier for that node (e.g., represented by integers in Fig. 1). Likewise, each connection gene indicates the identities of the two nodes that are being connected, the initial weight or strength (i.e., represented by the parameter $\omega$) of the connection and whether or not that connection has been enabled (i.e., set equal to a value of 0 or $\omega$), and a unique identifier for that connection (again, as represented by the integers in Fig. 1). The unique mutation identifiers provide a numerical index that can be used to reference each new innovation that occurs across successive generations. (In the exposition below, this index will be called the *innovation number*.)

---------------------------
Insert Figure 1 here

--------------------------

The complexity of the network topology increases over successive generations through genetic mutation. These mutations can affect both the number and type of nodes in a network, as well as the pattern of connectivity among the nodes. These two basic types of "connective" mutations are shown in Figure 2. New nodes are added to a network by replacing the connection between two existing nodes with an intermediary node that connects to the two original nodes (e.g., as shown in the left panel of Fig. 2). The new connections joining the new node to the original two are automatically enabled to ensure that the mutation affects the network's overall fitness. And similarly, mutations can affect the pattern of network connectivity by joining two previously unconnected nodes (e.g., as shown in the right panel of Fig. 2) or by changing the strength of an existing connection between two nodes. The probability that each type of mutation (i.e., adding a node, adding a connection, or modifying a connection) will occur is determined probabilistically, with the overall rates of each respective type of mutation being controlled by the parameters $\pi_{add\text{-}node}$, $\pi_{add\text{-}link}$, and $\pi_{mutate\text{-}link}$ (for a complete list of the parameter values associated with macroscopic evolution, see Appendix A).

The precise manner in which connections are modified also depends upon the "severity" of a mutation, which is determined probabilistically using the parameter $\delta_{severity}$. With probability $\delta_{severity}$, the mutation is considered to be "severe" and the connection is modified in one of the following mutually exclusive ways. With probability $c_{\text{cold-gauss}}$, the mutation in the existing connection is simply canceled, and with probability $1 - c_{gauss}$, the connection strength is sampled from a Gaussian distribution with $\mu_w = \omega$ (i.e., the initial weight value) and $\sigma_w = 0.5$; otherwise, the connection strength is sampled from a Gaussian distribution with $\mu_w = 0$ and $\sigma_w = 0.5$. On the other side, some connection genes, which are selected randomly, have probability $c_{\text{turn-on-off}}$ to change their connected condition (i.e., if the connection gene is disabled, it become enabled, vice versa). Finally, with the insertion of a new connection, there is some probability that two previously unconnected nodes cannot be found; when this happens, the parameter $\pi_{\text{attempt-mutation}}$ specifies the number of attempts that are made to locate the nodes before the effort is halted.

---------------------------
Insert Figure 2 here
---------------------------

The problem of aligning two genomes during crossover is computationally inexpensive because innovation numbers provide a basis for rapidly comparing any two genomes. Those genes that are identical between two parent genomes are said to *match*, and those that do not match are in either *disjoint* or *excess*, depending on whether they occur within or outside (respectively) of the range of innovation numbers of the other parent's genome. Using such information, crossover between the genomes of any two parents can occur in three different ways, as shown in Figure 3.

---------------------------
Insert Figure 3 here
---------------------------

In *single-point crossover*, an innovation number is randomly selected from the set of matching innovation numbers. All of the genes to the left of this point in one parent's genome (including disjoint genes but not excess genes because the latter by definition do not exist to the left of the crossover point) are then copied to the genome of the offspring, and all of the genes to the right of this point in the other parent's genome (including both disjoint and excess genes) are also copied to the genome of the offspring. If the gene at the crossover point happens to correspond to a connection weight, then that particular connection weight in the offspring is set equal to the mean of those connection weights in the two parents.

In *multipoint crossover*, each of the two possible values of all matching genes has an equal probability of being copied to the offspring's genome. And because all disjoint and excess genes can also be copied to the offspring's genome, the offspring's genome can become longer that the genome of either parent. One way to curtail this growth is to only copy disjoint and/or excess genes from the parent having the higher fitness value. With or without such precautions, however, this crossover method ensures that the offspring inherits both the shared and unique genes from both of its parents, thereby increasing both the combinatory and exploratory potential of this method relative to single-point crossover.

Finally, *multipoint-average crossover* is identical to multipoint crossover except that, rather than randomly assigning those matching genes representing the connection weights of one of the parents to the offspring, the offspring will instead inherit a gene for a connection weight that is the mean of the parents' two connection weights.

Each successive generation of networks is generated using some combination of the three aforementioned methods of crossover in conjunction with both cloning and mutation. More specifically, with each new generation, individual offspring can be produced in three ways, determined in a probabilistic manner: (1) via cloning or copying the genome of a network from one generation to the next; (2) via mutating a genome (using one of the methods described above) and then copying the mutated genome to the next generation; and (3) via one of the three crossover methods that were described above. As will be indicated below, the genome of the fittest individual in each species is simply cloned from one generation to the next; each remaining individual has a probability equal to $p_{mutate-only}$ of having a mutation introduced into its genome (as described above) and that genome being copied to the next generation, and a probability of 1 - $p_{mutate-only}$ of reproducing via crossover. With the latter, the method of crossover is determined probabilistically, with the probability of single-point, multipoint, and multipoint-average being specified by the parameters $p_{single-point}$, $p_{multipoint}$, and $p_{multipoint-average}$, respectively. Following crossover, the genome is left intact with probability $p_{mate-only}$ and a mutation is introduced to the genome with probability 1 – $p_{mate-only}$. Finally, although crossover typically occurs between two individuals from the same species, it can with some small probability, $c_{inter-species}$, occur between two individual from different species, sometimes resulting is offspring that are more fit than either parent.

    Using these evolutionary methods, populations of networks having complex and diverse topologies can evolve. However, because networks having simpler topologies tend to optimize more rapidly than networks having more complex topologies, the process of adding nodes and connections initially causes more

complex networks to be less fit. One method for preventing more complex network topologies from being prematurely removed from the population is to allow speciation, or the emergence of new "species" of networks (i.e., networks having more complex topologies) so that they can compete within their own more specialized ecological niches. By doing this, more complex networks are protected so that they have time to optimize their structures to their particular niches.

The process of speciation is also made computationally efficient by using the innovation numbers and by using the number of disjoint ($D$) and excess ($E$) genes as a metric of genome compatibility. The main intuition of this method is that pairs of genomes that contain large numbers of disjoint and/or excess genes are unlikely to share much of their evolutionary history, and are thus more likely to represent distinct species. This intuition is instantiated using a measure of the compatibility distance of two genomes. This distance, $\delta$, is a linear combination of $D$, $E$, and $\overline{W}$, or the mean weight differences of matching genes, and is specified by Equation 2.1:

$$(2.1) \quad \delta = \frac{c_1 E}{N} + \frac{c_2 D}{N} + c_3 \overline{W}$$

In Equation 2.1, $c_1$, $c_2$, and $c_3$ are coefficients to weight the relative contributions of the three relevant factors (i.e., excess genes, disjoint genes, and the mean weight difference of matching genes), and $N$ is the number of genes in the longer of the two genomes and is used to normalize genome length. Any two genomes having a compatibility distance exceeding some pre-specified threshold, $\delta_c$, are considered to belong to separate species and thus (usually) prohibited from breeding. In sorting genomes into species, each genome is grouped with the first species for which $\delta < \delta_c$, so that no genome belongs to more than one species.

Another potential problem associated with speciation is that one species can grow without bounds, taking over the entire population. To prevent this from happening, the number of individuals within a species is increased/decreased according to whether its ~~average~~ total fitness is above/below the mean fitness of the population of species. This is done using Equation 2.2:

$$(2.2) \quad N'_j = \frac{\sum_{i=1}^{N_j} f_{i,j}}{\bar{f}}$$

where $N_j'$ is the number of individuals in species $j$ adjusted for its mean fitness, as specified by the right side of the equation. There, $N_j$ is the non-adjusted number of individuals in species $j$, $f_{i,j}$ is the adjusted (relative) fitness of individual $i$ of species $j$ (as is given by Equation 2.3), and $\bar{f}$ is the mean fitness (i.e., the mean of $f_{i,j}$) of the entire population.

$$(2.3) \quad f_{i,j} = \frac{f_{i,j}^{raw} - f_{worst}}{\log(N_j)}$$

In Equation 2.3, $f_{i,j}^{raw}$ is the raw fitness of individual $i$ of species $j$ (which normal takes on a negative value) and $f_{worst}$ is the worst raw fitness in the population. Equation 2.3 differs from the original NEAT algorithm (Stanley & Miikkulainene, 2002) in that the denominator is the logarithm of $N_j$ rather than $N_j$ to take advantage of fitter species. The size of each species is thus adjusted so that the number of individuals in more fit species tends to increase and the number of individuals in less fit species tends to decline. With each new generation, a certain percentage (determined by the parameter $c_{survival}$) of the best performing (i.e., most fit) of each species is allowed to randomly produce the next generation of for their species.

A fitness amplification assumption is also added to the original NEAT algorithm that allows the fittest individual in the population to have the unique opportunity to contribute a copy of its genome (i.e., a clone of itself) to the next generation. This is done using Equation 2.4, where $f_{best}$ is the adjusted fitness value of the best-performing member of a species, and $c_{best}$ is an amplification coefficient that enhances that fitness value so as to ensure that that individual contributes a copy of itself to the next generation.

(2.4) $\quad f_{best} \leftarrow c_{best} f_{best}$

To further enhance the reproductive advantage of better species and thereby increase the probability of evolution being successful, a "delta coding" procedure is introduced that "steals" or eliminates a certain number of offspring, $d_{offspring\text{-}stolen}$, from whatever species has improved the least. In a similar manner, when the least fit species has not improved over a certain number of generations, $d_{drop\text{-}off\text{-}age}$, its overall fitness is reduced; to ~~product~~ protect younger species, their fitness is increased by an amplification factor, $d_{age\text{-}signifance}$.

It is worth emphasizing that the algorithm as described so far is biased to favor networks having simple topologies. The reason for this is that the addition of nodes and/or connections is not without cost: As the topology of a network increases, so too does the inherent difficult associated with evaluating its contribution to a network's fitness. Thus, by starting the evolutionary process with the simplest possible networks (i.e., networks without hidden units), more complex topologies are only retained if they are justified on the grounds that they increase a network's fitness.

Finally, the probability of mutation is determined by a *simulated annealing* process (Černý, 1985; Kirkpatrick, Gelatt, & Vecchi, 1983) in which the overall probability is initial some large value but then declines with each successive generation. (The "simulated annealing" metaphor comes from metallurgy, were the initial high temperature of a metal is slow decreased over time so that the atoms can align in a way that ensures high tensile strength.) The rate of this decline is controlled by a "temperature" parameter, $T$ (i.e., to simplify, $T$ is set equal to generations from last reset), and the probabilities of adding either a new node, $\pi_{add\text{-}node}$, or a new connection link, $\pi_{add\text{-}link}$, to a network are given by Equations 2.5 and 2.6, respectively:

(2.5) $\pi_{add-node} = \max[\psi_1, \min(\psi_2, x)]$, where $x \leftarrow x - \dfrac{1}{k_1 T}$

(2.6) $\pi_{add-link} = \max[\psi_3, \min(\psi_4, x)]$, where $x \leftarrow x - \dfrac{1}{k_2 T}$

where $\psi_1$ and $\psi_2$ respectively represent the lower and upper limits for the probability of adding a node via mutation, $\psi_3$ and $\psi_4$ respectively represent the lower and upper limits for the probability of adding a link via mutation, and $k_1$ and $k_2$ represent coefficients for adjusting the probabilities of adding nodes and links, respectively. Similarly, in Equation 2.7, $\psi_5$ and $\psi_6$ respectively represent the lower and upper limits for the probability of reproduction via mutation only, $\Delta$ is a parameter that control how much the probability is incremented per generation, and $k_3$ is a dynamic coefficient that is set equal to -1 if the fitness of the population fails to improve by some criterion, $c_{annealing}$ (otherwise, $k_3$ is set equal to 1). Finally, if the fitness of the population fails to improve, the values of the $\pi_{add\text{-}node}$ and $\pi_{add\text{-}link}$ are set equal to the starting values.

(2.7) $\quad p_{mutate-only} = \max[\psi_5, \min(\psi_6, x)]$, where $x \leftarrow x + k_3\Delta$

In the next section of this article, we explain how the process of microscopic evolution or learning also contributes to network topology and how this, in turn, determines the fitness of the network and hence its likelihood of contributing to the gene pool of the next generation.

## 3. Microscopic Evolution (or Learning) of the Network Structure

Our algorithm for microscopic evolution of network connection weights is based on the *Covariance Matrix Adaptation Evolution Strategy* (*CMA-ES*; Hansen, 2006; Hansen & Kern, 2004; Hansen, Müller, & Koumoutsakos, 2003; Hansen & Ostermeier, 2001; Iger, Hansen, & Roth, 2007; Suttorp, Hansen, & Igel, 2009). This algorithm provides a stochastic method for parameter optimization of non-linear, non-convex functions. As such, it is particularly useful for "rugged" parameter landscapes comprised of discontinuities and local optima (e.g., sharp "ridges") and is thus well suited to solve ill-conditioned and non-separable problems. Some modifications of the CMA-ES algorithm were necessary, however, to make it more amenable to the problem of evolving network connection weights.

The general intuition behind the algorithm is that, rather than training a network across a series of trials to find the set of connection weights that allow a network to solve some problem, the connection weights are instead evolved. To understand how this is done, it is first necessary to understand that the weights themselves can be represented as a vector, and that the elements of this vector can be sampled to find values that allow a network to solve a particular problem. In essence, this is what the CMA-ES algorithm does: During each learning trial, a set of vectors representing possible connection weight solutions to the problem are sampled from a sampling distribution, and then a new mean and covariance of a sampling distribution are computed that reflect the fitness of these sampled vectors. This whole process is repeated until either a solution meeting some goodness-of-solution criterion has been reached, or a stopping criterion has been reached.

During each trial $t+1$, the $\lambda$ individual vectors (i.e., candidate solutions consisting the connection weights) are sampled using Equation 3.1:

(3.1) $\quad x_k^{t+1} \sim N\left(\langle x \rangle_w^{(t)}, \sigma^{2(t)} C^{(t)}\right) k = 1,...\lambda$

where $N\left(\langle x \rangle_w^{(t)}, \sigma^{2(t)} C^{(t)}\right)$ is a normally distributed vector with mean $\langle x \rangle_w^{(t)}$, sampling variance $\sigma^{2(t)}$, and covariance matrix $C^{(t)}$. (For a complete list of the parameter values associated with microscopic evolution, see Appendix B.) This provides a simple method for sampling candidate vectors of connection weights using a multi-normal distribution, with the covariance matrix determining the degree to which sampling proceeds in a cautious versus audacious manner. The mean of the sampling distribution is computed using the weighted average of the individual sampled vectors using Equation 3.2:

(3.2) $\quad \langle x \rangle_x^{(t)} = \sum_{i=1}^{\mu} w_i x_{i:\lambda}^{(t)}$

with the constraints that $w_i > 0$ for all values of $i$ and $\sum_{i=1}^{\mu} w_i = 1$, and with the index $i{:}\lambda$ denoting the $i$-th best individual. By combining the best connection-weight vectors in this weighted manner, the mean of the sampling distribution approaches the target solution over time.

In Equation 3.1, $C^{(t)}$, represents the correlation between different connection weights within a neural network, and $\sigma^{2(t)}$ modulates the variability associated with the sampling distribution. Both terms are therefore vital to the success of the algorithm. Over trials, $C^{(t)}$ changes as a function of both the *evolutionary path*, which as its name suggests, controls the global orientation or trajectory of the evolutionary process, thereby allowing it to converge towards a solution. $C^{(t)}$ is updated using Equation 3.3, were $c_{cov}$ is a parameter that controls the rate of change, $\mu_{cov}$ is a parameter for weighting between the rank-one and rank-$\mu$ update, and the terms $P_c^{(t+1)}$ and $S^{(t+1)}$ are specified by Equations 3.4 and 3.5, respectively.

(3.3) $\quad C^{(t+1)} = (1 - c_{cov})C^{(t)} + c_{cov}\dfrac{1}{\mu_{cov}}P_c^{(t+1)} + c_{cov}\left(1 - \dfrac{1}{\mu_{cov}}\right)S^{(t+1)}$

(3.4) $\quad P_c^{(t+1)} = p_c^{(t+1)}\left(p_c^{(t+1)}\right)^T$

(3.5) $\quad S^{(t+1)} = \sum_{i=1}^{\mu} \dfrac{w_i}{\sigma^{2(t)}}\left(x_{i:\lambda}^{(t+1)} - \langle x \rangle_w^{(t)}\right)\left(x_{i:\lambda}^{(t+1)} - \langle x \rangle_w^{(t+1)}\right)^T$

In Equation 3.4, the term $p_c^{(t+1)}$ is specified by Equation 3.6, which accumulates the differences between the mean connection-weight sampling-distribution vectors across successive trials:

(3.6) $\quad p_c^{(t+1)} = (1 - c_c)p_c^{(t)} + H_\sigma^{(t+1)}\sqrt{c_c(2 - c_c)}\dfrac{\sqrt{\mu_{eff}}}{\sigma^{(t)}}D^{(t+1)}$

where $c_c$ is learning rate for accumulation for the rank-one update of the covariance matrix, and $H_\sigma^{(t+1)} = 1$ if $\dfrac{\|p_\sigma^{(t+1)}\|}{\sqrt{1-(1-c_\sigma)^{2\tau}}} < \left(1.4 + \dfrac{2}{n+1}\right) E(\|N(0,I)\|)$; otherwise, $H_\sigma^{(t+1)} = 0$.

(In the preceding conditional statement, the index $\tau$ denotes the number of completed trails.) The term $\mu_{eff}$ is the *variance effective selection mass*, which is related to recombination weights and is constrained so that $\mu_{eff} = 1 \big/ \sum_{i=1}^{\mu} w_i^2$. Finally, the last term in Equation 3.6 is the actual difference between mean sampling distribution vectors across successive trials and is specified by Equation 3.7:

(3.7) $\quad D^{(t+1)} = \left(\langle x \rangle_w^{(t+1)} - \langle x \rangle_w^{(t)}\right)$

Finally, the sampling variance in Equation 3.1, $\sigma^{2(t)}$, controls the overall rate of change in the evolutionary process. The value of $\sigma^{2(t)}$ during any given trial is given by Equation 3.8, in which the term $p_\sigma^{(t+1)}$ is specified by Equation 3.9.

(3.8) $\quad \sigma^{(t+1)} = \sigma^{(t)} \exp\left[\dfrac{c_\sigma}{d_\sigma}\left(\dfrac{\|p_\sigma^{(t+1)}\|}{E(\|N(0,I)\|)} - 1\right)\right]$

(3.9) $\quad p_\sigma^{(t+1)} = (1-c_\sigma)p_\sigma^t + \sqrt{c_\sigma(2-c_\sigma)} B^{(t)} E^{-1(t)} B^{T(t)}$

where the orthogonal matrix, $B^{(t)}$, and the diagonal matrix, $E^{(t)}$, are obtained via principal component analysis of $C^{(t)}$ using the matrix theorem: $C^{(t)} = B^{(t)} E^{2(t)} B^{T(t)}$. In Equation 3.8, $E(\|N(0,I)\|)$ is the expected length of $p_\sigma$ under random selection and is given by Equation 3.10:

$$(3.10) \quad E(\|N(0,I)\|) = \frac{\sqrt{2}\Gamma\left(\frac{n+1}{2}\right)}{\Gamma\left(\frac{n}{2}\right)} \approx \sqrt{n}\left(1 - \frac{1}{4n} + \frac{1}{21n^2}\right)$$

where *n* denotes the number of search-space dimensions, which in this application simply corresponds to the number of connection weights.

As indicated above, our algorithm modifies the standard CMA-ES algorithm (as described so far) to automatically inflate the population size, $\lambda$, and the number of trial iterations for the stopping condition, $\tau_{stop}$, whenever the algorithm *stagnates*, or stops converging towards a better solution. This is done using Equations 3.11 and 3.12, respectively, where *o* denotes the number of generations over which the algorithm has stagnated, and $\rho$ is a parameter that denotes the default number of generations.

$$(3.11) \quad \lambda = 4 + \lfloor 3\ln(n+o) \rfloor$$

$$(3.12) \quad \tau_{stop} = \rho(1+o)$$

Furthermore, to improve microscopic evolution, especially when the population is trapped in local maxima, the sampling variance in initial trial, $\sigma^{(0)}$, is defined according to its stagnation level using Equation 3.13. By increasing the sample, the population may thereby escape the local maxima point.

$$(3.13) \quad \sigma^{(0)} = \min(\sigma_{max}, \sigma_d + o_\sigma \cdot \log(o+1))$$

where $\sigma_{max}$ represents the upper limit of initial sampling variance, $\sigma_d$ represents the default sampling variance and $o_\sigma$ represents coefficient for scaling stagnation effect.

Finally, a fitness criterion can also be used to stop the evolutionary process. That is, if the fitness of the network, $\vartheta$, exceeds some fitness threshold, $f_{stop}$, or the trial iteration exceeds the stop condition, $\tau_{stop}$, the microscopic evolution of the connection weights is halted.

## 4. Reinforcement Learning of Connection Weights

The previous two sections have described how network topology is generated through the processes of macro- and microscopic evolution. In this final section, we describe the algorithm that is used to train a network based on reinforcement learning. As indicated, reinforcement learning refers to a general class of machine-learning algorithms in which performance is "shaped" using a single training signal corresponding to the reward/punishment that is associated with specific actions and/or the states that result from those actions (Sutton & Barto, 1998). Of central importance to this notion is the idea of *reward prediction error*, usually denoted by $\delta$, which represents the reward that an artificial agent anticipates in response to a particular action and the state that then results from that action. This is represented in Equation 4.1, in which *R* represents the immediate reward that is received from executing the action, *x* represents a particular state that the agent can be in at time *t*, and the *value function V* represents the value associated with the state. The parameter $\gamma$ is a *discount* parameter that determine how much the reward that is anticipated to result from the next state, $x_{t+1}$, is weighed against the immediate reward; small values of $\gamma$ thus make the agent "greedy" in that it tends to prefer actions that result in large immediate rewards, whereas large values of $\gamma$ cause the agents to prefer actions that result in rewards over the long run. (For a complete list of the parameter values associated with reinforcement learning, see Appendix C.)

(4.1) $\delta = R + \gamma V(x_{t+1}) - V(x_t)$

One way to implement this algorithm within a neural network is to use the standard error-driven back-propagation algorithm (Rumelhart et al., 1986), training the network using each state at time $t$, $x_t$, as the input, and allowing the resulting state at time $t+1$, $x_{t+1}$, as the output. The teaching signal for the desire output is then $R + \gamma V(x_{t+1})$, and the weight for any given connection in the network is adjusted during learning using Equation 4.2:

(4.2) $\Delta w_{direct} = \alpha [R + \gamma V(x_{t+1}) - V(x_t)] \dfrac{\partial V(x_t)}{\partial w}$

where $\alpha$ is a learning rate parameter that controls the rate of convergence and the subscript "direct" denotes the name of this algorithm. Although this *direct algorithm* has been used successfully in many applications (Tesauro, 1990, 1992), it is not guaranteed to converge for general function-approximation systems (Baird, 1995).

To develop such an algorithm, the problem can be restated as being one of predicting the outcome of a deterministic Markov chain, with the goal being to specify a value function that, for any given state, $x_t$, will give the value of the immediate reward and the successor state, $x_{t+1}$, thereby satisfying the *Bellman equation* (Bellman, 1957):

(4.3) $V(x_t) = \langle R + \gamma V(x_t) \rangle$

where $\langle \ \rangle$ is the expected value of all possible successor states, $x_{t+1}$. For a system having a finite number of states, the optimal value function, $V^*$, will provide a unique solution to the above equation, and any value function that is suboptimal will result in an inequality called the *Bellman residual*. For a system with $n$ states, the mean squared Bellman residual is given by Equation 4.4, and provides a direct measure, $E$, of the degree to which a given policy is suboptimal. And because the value of $E$ is bounded, it suggests an alternative to the direct algorithm for adjusting the connection weights in a neural network function-approximation system: by performing stochastic gradient descent on $E$.

(4.4) $$E = \frac{1}{n}\sum_x \langle R + \gamma V(x_{t+1}) - V(x_t) \rangle^2$$

Under the assumption that $V$ is parameterized by the set of connection weights, the adjustment to any given weight $w$ following a transition from $x_t$ to $x_{t+1}$ with reward $R$ is specified by Equation 4.5:

(4.5) $$\Delta w_{residual-gradient} = -\alpha [R + \gamma V(x_{t+1}) - V(x_t)] \left[ \gamma \frac{\partial}{\partial w} V(x_{t+1}) - \frac{\partial}{\partial w} V(x_t) \right]$$

where the "residual-gradient" subscript denotes the name of the algorithm, *residual-gradient*. For a system with a finite number of states, $E$ will equal 0 only if $V = V^*$. And critically, this residual-gradient algorithm is guaranteed to converge, thus making it ideally suited for training neural networks to be function-approximation systems. The one limitation of this algorithm, however, is that it is slow (Baird, 1995; Williams & Baird, 1993). This limitation makes the algorithm impractical for large-scale problems of the type that might be of interest to psychologists (e.g., Reichle & Laurent, 2006). This limitation results in the following quandary: Whereas the direct algorithm is rapid, it is not guaranteed to converge, especially for large problems; in contrast, the residual-gradient algorithm is guaranteed to converge, but is slow, especially for large problems. Because both algorithms are based on gradient descent, however, the solution to this problem is fairly straightforward: One simply combines the two algorithms so that the adjustment to any given connection weight *w* is simply some weighted average of the adjustments given by the direct and residual-gradient algorithms. This is done using Equation 4.6:

(4.6) $\Delta w_{residual} = (1-\phi)\Delta w_{direct} + \phi \Delta w_{residual-gradient}$

where the subscript "residual" denotes that this is the *residual algorithm*. In the equation, the parameter $\phi$ modulates the degree to which the direct and residual-gradient algorithms contribute to the adjustment of a given weight, *w*. Our algorithm thus benefits from the strengths of the direct and residual-gradient algorithms by combining the speed of the former with the convergence of the latter. In the next section, we will indicate how this residual algorithm is used in conjunction with our algorithms for macro- and microscopic evolution to enable artificial neural networks that are capable of rapidly learning very large and complex problems.

**5. Combining Macro- and Microscopic Evolution with the Residual Algorithm**

Figure 4 is a schematic diagram illustrating how our algorithms for implementing macro- and microscopic evolution to generate artificial network topologies are combined with the residual algorithm to generate networks that are capable of solving large, complex problems via error-driven reinforcement learning. This process starts by first generating a population of simple genomes that express themselves as individual networks (i.e., as phenotypes). Each network is then trained on the same problem using the residual gradient to adjust the network's connection weights. After training, each network's performance is then evaluated by summing the reward that it received (i.e., this is the measure of a network's fitness). If the overall fitness of the population improves, then the individual networks are allowed to generate the next generation via cloning the fittest individual, and via mutation and crossover. If, however, the fitness does not improve (i.e., the generation stagnates), then the microscopic evolutionary algorithm is used. (Because the latter is computationally expensive, it is only used as necessary, to "nudge" the evolutionary process towards a better solution.)

---------------------------
Insert Figure 4 here
---------------------------

## 6. General Discussion

This article has described a procedure whereby macro- and microscopic evolutionary algorithms are jointly used to specify the topology of artificial neural networks that are capable of learning complex, large-scale problems via error-driven reinforcement learning. In this final section, we will briefly consider both the applied and theoretical implications of this new approach to simulating learning.

The main practical benefit of our approach is that is provides a novel way to solve problems that have here-to-fore either been too large or complex to effectively solve using standard reinforcement-learning algorithms. To cite just one example, the algorithm has been used to simulate and understand the complex patterns of eye movements that are observed when people read text (Liu & Reichle, 2010; Reichle, Liu, & Laurent, 2011). The artificial networks in these simulations were generated as described in this article and were given the task of learning to move their "eyes" and "attention" so at to "read" as efficiently as possible (i.e., identify sequences of virtual "words" in the minimal amount of time). The networks were rewarded for each word it identified, and punished for each time step that it spent processing a given sentence. With these incentives, the agents learned to direct their eyes towards the centers of words because, with limited visual acuity, those locations afforded to most rapid identification of words. The agents also learned the modest relationship between word length and word-processing difficulty (i.e., longer words on average required more time to identify) and then learned to exploit this knowledge by initiating saccadic programming (which required some amount of time to complete) so that the eyes would move off of a word just as it was being identified. This "strategy" for deciding where and when to move the eyes was optimal in that it allowed the networks to fixate each word only as long as necessary; initiating saccadic programming any sooner would cause the eyes to move too soon, whereas initiating programming any later would cause the fixations to be unnecessarily long in duration. Importantly, these simulations replicated earlier results that had been completed using a reinforcement-learning algorithm (i.e., *value-iteration*) that had been instantiated using a look-up table (Reichle & Laurent, 2006), but extended these results by using more complex and realistic assumptions that would not have been practically feasible using a look-up table. These simulations also extended the earlier ones by showing that the agents were able to generalize their eye-movement behaviors to novel stimuli (i.e., sentences and words that had not been encountered during training). We suspect that the capacity to simulate complex cognitive tasks and to generalize behavior to novel experiences will prove to be one of the main benefits of error-driven reinforcement learning. Returning to the example of eye-movement control in reading, the networks are currently being used (Liu, 2011) to examine the issue of how attention is allocated during reading—to one word at a time, in parallel to multiple words, or via some more complex, dynamic allocation scheme.

      From a theoretical perspective, the main contribution of our approach is that is suggests the important role that evolution might play in generating network topologies that are capable of integrating the principles of error-driven and reinforcement learning. The reason for this is that the network topology is an important determinant of the types of problems a network is able to learn; by evolving topologies capable of implementing error-driven learning through reinforcement, our networks are able to exploit the function-approximation capabilities of artificial neural networks using a one-dimension error signal. This effectively provides a way to exploit the strengths of artificial neural networks while sidestepping one of their main practical and theoretical limitations—the requirement to provide an explicit error (i.e., "teaching") signal to the networks during their training. By instantiating error-driven reinforcement learning, our networks are capable of solving the types of large-scale and complex problems that biological organisms face in their struggle to survive and reproduce, and that are of interest to researchers in various problem domains (e.g., engineering; Sejnowski, 2010).

      Finally, while both the macroscopic evolutionary and reinforcement learning components of our approach have obvious direct analogs in biology and psychology,

respectively, one might wonder about the possible analog of our algorithm for microscopic evolution. As it turns out, the central assumption of this algorithm—that network connection weights are sampled from a distribution that is updated across time to reflect how well the sampled weights solve a particular problem of interest—may have a direct analog in the spontaneous oscillation that is very commonly observed in many neuronal systems (Sejnowski, 2006). Such oscillation is known to occur during sleep (Sejnowski, 2000; Takashima, 2006) and has been posited to play a significant functional role in memory consolidation by organizing pre- and post-synaptic spike times, thereby facilitating spike-timing-dependent plasticity (Paulsen & Sejnowski, 2000). Although it is not known exactly how these synaptic weights changes occur, spontaneous oscillation may contribute to optimal network performance by, for example, reducing the dimensionality of the weight space to permit more efficient learning (Montague & Sejnowski, 1994; Rao & Sejnowski, 2003). Thus, on some abstract level, our microscopic evolution algorithm might be analogous to the spontaneous oscillation that has been observed in real networks and that has been posited to play an important optimization role. Future work will be necessary, however, to validate this potential correspondence.

Thorndike, E. L. (1901). Animal intelligence: An experimental study of the associative processes in animals. *Psychological Review Monograph Supplement, 2*, 1-109.

Williams, R. J. and Baird, L. C. (1993). Tight performance bounds on greedy policies based on imperfect value functions. *Northeastern University Technical Report NU-CCS-93-14, November.*

Ye, C., Yung, N.-C., & Wang D. (2003). A fuzzy controller with supervised learning assisted reinforcement learing algorithm for obstacle avoidance. *IEEE Transctions on Systems, Man, and Cybernectics – PART B: Cybernetics,* 33, 17-27.
36

**8. Acknowledgements**

Correspondence regarding this article should be addressed to Erik Reichle, University of Pittsburgh, 635 LRDC, 3939 O'Hara St., Pittsburgh, PA, 15260; or via e-mail to: reichle@pitt.edu.  The work described in this article was supported by an award from the Chinese Scholarship Council to the first author and an NIH R01 grant (HD053639) that was awarded to the second author.  The source code for both evolutionary algorithm and error-driven reinforcement learning networks is available upon request from the first author.

## 9. Appendix A

This appendix contains tables listing the parameters for the macroscopic evolution algorithm (see Section 2), their functional roles, and their default values.

*Table A.1*: Parameters controlling mutation.

| Parameter | Description | Value |
|---|---|---|
| $c_{cold\text{-}gauss}$ | probability of canceling weights mutation in existing connection | 0.1, if $\delta_{severity}$ = true; 0.3, otherwise |
| $c_{gauss}$ | = 1 – probability of sampling new connection from distribution having the same mean connection strength | 0.3, if $\delta_{severity}$ = true; 0.5, otherwise |
| $C_{turn\text{-}on\text{-}off}$ | probability of turning on or off the selected connection gene | 0.2 |
| $\delta_{severity}$ | probability of "severe" connection weight mutation | 0.5 |
| $\pi_{add\text{-}link}$ | Probability of adding connection | 0.6 |
| $\pi_{add\text{-}node}$ | Probability of adding node | 0.2 |
| $\pi_{attempt\text{-}mutation}$ | Number of attempts to locate "lost" nodes | 50 |
| $\pi_{mutate\text{-}link}$ | Probability of connection mutation | 0.9 |
| $\sigma_w$ | Standard deviation of mutated connection distribution | 0.5 |

*Table A.2*: Parameters controlling the reproduction of offspring.

| Parameter | Description | Value |
|---|---|---|
| $c_{best}$ | fitness amplification coefficient for best genome | 3 |
| $c_{inter\text{-}species}$ | probability of inter-species mating | 0.2 |
| $c_{survival}$ | percent of genome per species that survives | 0.2 |
| $p$ | population size | 120 |
| $p_{mate\text{-}only}$ | probability of reproducing only via mating | 0.2 |
| $p_{multipoint}$ | probability of multipoint crossover | 0.6 |
| $p_{multipoint\text{-}average}$ | probability of multipoint-average crossover | 0.4 |
| $p_{mutate\text{-}only}$ | probability of mutating only | 0.3 |
| $p_{single\text{-}point}$ | probability of single-point crossover | 0.3 |

*Table A.3*: Parameters controlling the compatibility distance metric.

| Parameter | Description | Value |
|---|---|---|
| $c_1$ | scaling factor for disjoint genes | 1 |
| $c_2$ | scaling factor for excess genes | 1 |
| $c_3$ | scaling factor for gene differences | 2 |
| $\delta_c$ | minimal distance for 2 genomes to be in same species | 3 |

*Table A.4*: Parameters controlling the "delta coding" procedure.

| Parameter | Description | Value |
|---|---|---|
| $d_{age\text{-}significance}$ | age amplification factor | 1 |
| $d_{drop\text{-}off\text{-}age}$ | generations after which poor species are penalized | 2000 |
| $d_{offspring\text{-}stolen}$ | number of offspring eliminated from unfit species | 10 |

Table A.5: Parameters controlling simulated annealing.

| Parameter | Description | Value |
|---|---|---|
| $C_{annealing}$ | criterion for simulated annealing | 10 |
| $\Delta$ | increment for adjusting $p_{mutate\text{-}only}$ | 0.01 |
| $k_1$ | coefficient for adjusting probability of adding node | 20 |
| $k_2$ | coefficient for adjusting probability of adding link | 10 |
| $\psi_1$ | lower probability limit of adding node | 0.02 |
| $\psi_2$ | upper probability limit of adding node | 0.04 |
| $\psi_3$ | lower probability limit of adding link | 0.1 |
| $\psi_4$ | upper probability limit of adding link | 0.2 |
| $\psi_5$ | lower limit of $p_{mutate\text{-}only}$ | 0.3 |
| $\psi_6$ | upper limit of $p_{mutate\text{-}only}$ | 0.5 |

## 10. Appendix B

This appendix contains a table listing the parameters for the microscopic evolution algorithm (see Section 3), their functional roles, and their default values. (Note that $n$ represents the number of network connections and $o$ represents the number of generations that have stagnated.)

*Table B.1*: Parameters controlling the initial settings.

| Parameter | Description | Value |
|---|---|---|
| $C^{(0)}$ | covariance matrix | $I$ (i.e., identity matrix) |
| $p_c^{(0)}$ | evolutionary path for covariance matrix | 0 |
| $p_\sigma^{(0)}$ | evolutionary path for variance | 0 |
| $\sigma_{max}$ | upper limit for sample variance | 1.0 |
| $\sigma_d$ | default value for sample variance | 0.5 |
| $o_\sigma$ | coefficient for scaling stagnation effect | 0.01 |

*Table B.2*: Parameters controlling the Covariance Matrix Adaptation.

| Parameter | Description | Value |
|---|---|---|
| $c_c$ | learning rate for accumulation for rank-1 update of the covariance matrix | $4 / (n+4)$ |
| $c_{cov}$ | initial learning rate for covariance matrix update | $\frac{1}{\mu_{cov}} \cdot \frac{2}{(n+\sqrt{2})} + \left(1 - \frac{1}{\mu_{cov}}\right) \min\left(1, \frac{2\mu_{eff} - 1}{(n+2)^2 + \mu_{eff}}\right)$ |
| $\mu_{cov}$ | weighting between rank-1 and rank-$\mu$ update | $\mu_{eff}$ |

*Table B.3*: Parameters controlling the step size.

| Parameter | Description | Value |
|---|---|---|
| $c_\sigma$ | learning rate for accumulation of step-size control | $\dfrac{\mu_{eff} + 2}{n + \mu_{eff} + 3}$ |

| Parameter | Description | Value |
|---|---|---|
| $d_\sigma$ | damping for step-size update | $1 + 2\max\left(0, \sqrt{\dfrac{\mu_{eff}-1}{n+1}} - 1\right) + c_\sigma$ |

*Table B.4*: Parameters controlling selection and recombination.

| Parameter | Description | Value |
|---|---|---|
| $\lambda$ | candidate sample size | $4+[3\ln(n+o)]$ |
| $\mu$ | parent size | $\lambda/2$ |
| $\mu_{eff}$ | variance effective selection mass | $\dfrac{1}{\sum_{i=1}^{\mu} w_i^2}$ |
| $w_{i=1\ldots\mu}$ | recombination weights | $\dfrac{\ln(\mu+1)-\ln(i)}{\sum_{j=1}^{\mu}[\ln(\mu+1)-\ln(j)]}$ |

*Table B.5*: Parameters controlling the stopping conditions.

| Parameter | Description | Value |
|---|---|---|
| $f_{stop}$ | fitness stopping criterion | best fitness in history |
| $\rho$ | default stopping coefficient | 1000 |

# 11. Appendix C

This appendix describes default parameters for reinforcement learning of connection weights (section 4).

*Table C.1*: Initial parameters for the reinforcement learning of network connection weights.

| Parameter | Description | Value |
|---|---|---|
| $\alpha$ | Learning rate | 0.05 |
| $\gamma$ | Discount rate | 0.9 |
| $\phi$ | Balance rate | 0.5 |

# 11. Figure Captions

*Figure 1.* An example showing the coding and mapping of a single genotype (in the top panel) onto its corresponding phenotype, an artificial neural network (in the bottom panel). As indicated, the genotype codes node identity and type, as well as information about connections between nodes, including their pattern of connectivity, weight, innovation number, and whether or not they are enabled.

*Figure 2.* An example illustrating two types of mutations: On the left, the mutation results in the addition of a node, while on the right, the mutation results in the addition of a connection weight between two nodes.

*Figure 3.* The top panel shows the genomes of two parents, along with the corresponding networks. The middle panel shows the two genomes aligned, as would occur during crossover. The bottom panel shows the three different crossover methods (single-point, multipoint, and multipoint-average) and the resulting offspring phenotypes.

*Figure 4.* A schematic diagram of the evolutionary process, in its entirety. The top box shows the population, which consists of *N* species of genotypes and their resulting phenotype networks. Each generation is trained on a problem using the residual-gradient reinforcement-learning algorithm. If the overall fitness of that generation improves (i.e., it does not stagnate), production procedure are allowed to occur directly (mutation or crossover or clone depended on random probability); otherwise, the CMA-ES algorithm is used to sample network connection weights and thereby produce the next generation.

Figure 1.

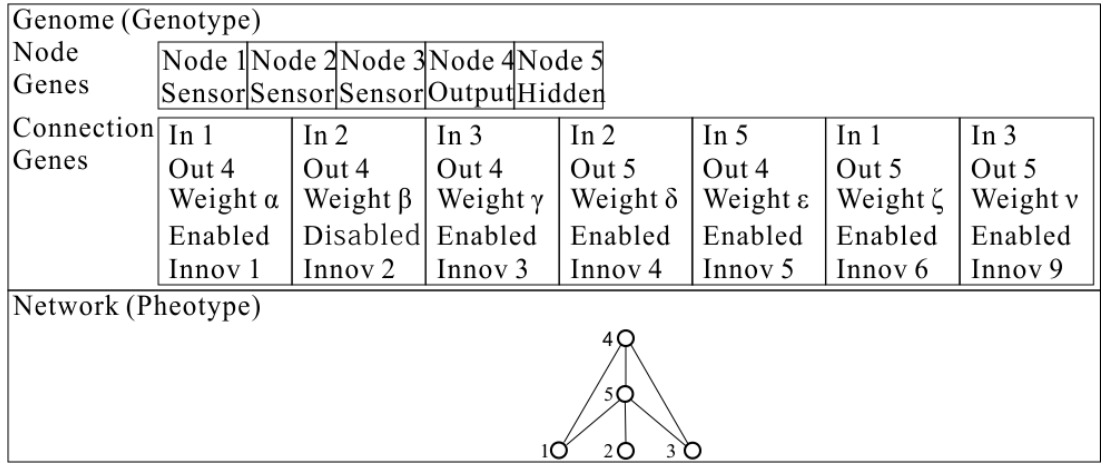

Figure 2.

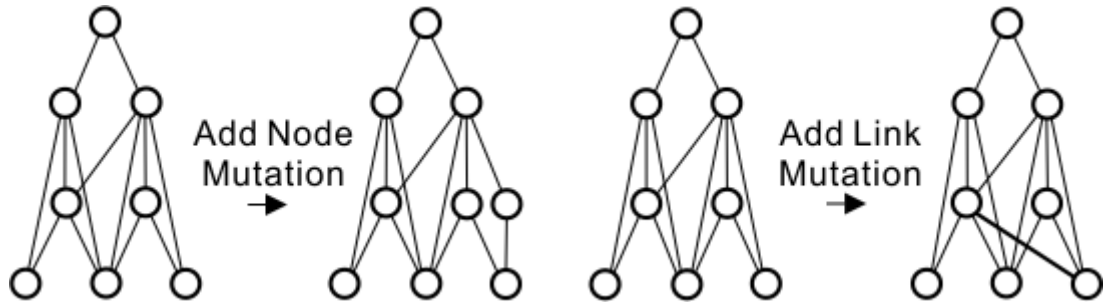

Figure 3.

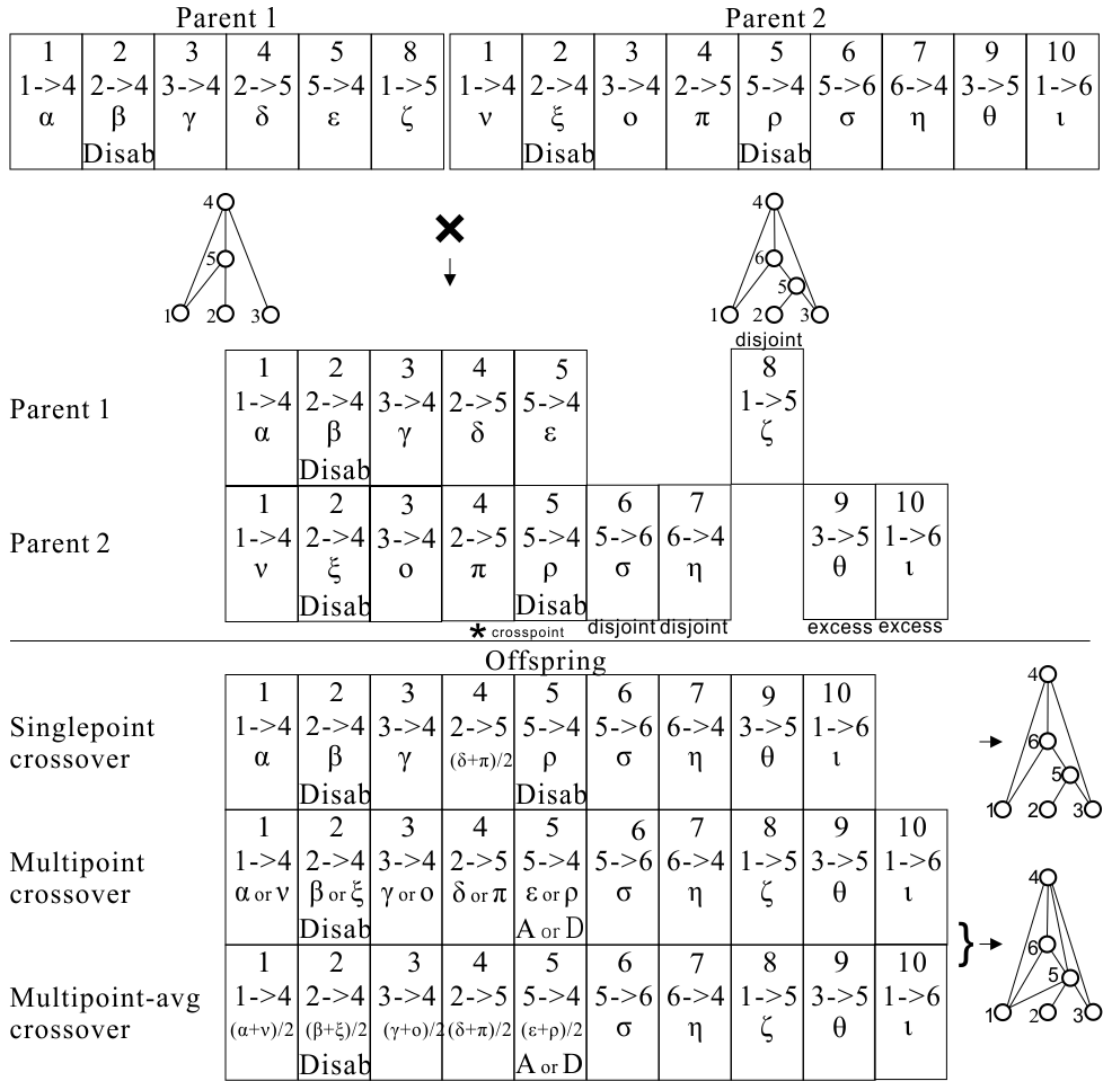

Figure 4.

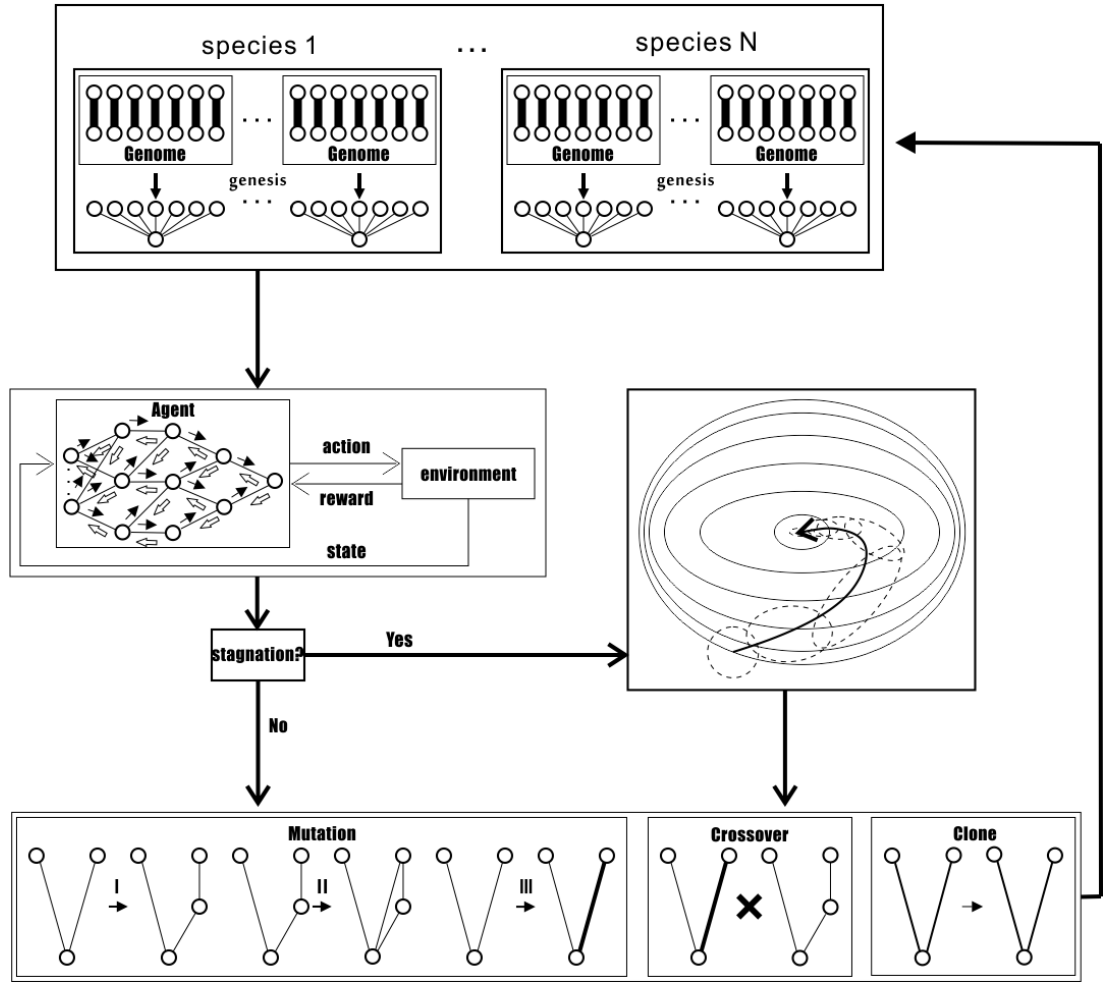